\documentclass[12pt]{article}

\usepackage{natbib}

\usepackage{graphicx}
\usepackage{hyperref}
\usepackage{url}

\usepackage{subfigure}
\usepackage{algorithm}
\usepackage{algorithmic}

\usepackage{amsfonts}
\usepackage{amsmath}
\usepackage{amsbsy}
\usepackage{multirow}
\usepackage{booktabs}

\newcommand{\captionfonts}{\normalsize}

\makeatletter
\long\def\@makecaption#1#2{%
  \vskip\abovecaptionskip
  \sbox\@tempboxa{{\captionfonts #1: #2}}%
  \ifdim \wd\@tempboxa >\hsize
    {\captionfonts #1: #2\par}
  \else
    \hbox to\hsize{\hfil\box\@tempboxa\hfil}%
  \fi
  \vskip\belowcaptionskip}
\makeatother

\newcommand{\ie}{{i.e.}~}
\newcommand{\eg}{{e.g.}~}
\newcommand{\wrt}{{w.r.t.}~}

\newcommand{\E}{\textrm{E}}
\newcommand{\EE}{\displaystyle \mathop{\mathbb{E}}}
\newcommand{\mytilde}{\raise.17ex\hbox{$\scriptstyle\mathtt{\sim}$}}
\newcommand{\Erbm}{\E}
\newcommand{\Eorbm}{\E}
\newcommand{\Prbm}{P}
\newcommand{\Porbm}{P}
\newcommand{\Pirbm}{P}
\newcommand{\Frbm}{F}
\newcommand{\Forbm}{F}
\newcommand{\Firbm}{F}

\newcommand{\cD}{\mathcal{D}}
\newcommand{\cS}{\mathcal{S}}
\newcommand{\cV}{\mathcal{V}}
\newcommand{\cH}{\mathcal{H}}
\newcommand{\BigO}{O}

\newcommand{\sigm}{\sigma}
\newcommand{\bsigm}{\boldsymbol\sigma}

\newcommand{\softplus}{\textrm{soft}_+}

\newcommand{\ind}{\mathbf{1}}

\newcommand{\bv}{\mathbf{v}}

\newcommand{\bW}{\mathbf{W}}
\newcommand{\bh}{\mathbf{h}}
\newcommand{\bbvis}{{\mathbf{b}^\text{v}}}
\newcommand{\bbhid}{{\mathbf{b}^\text{h}}}
\newcommand{\bvis}{b^\text{v}}
\newcommand{\bhid}{b^\text{h}}

\DeclareMathOperator*{\argmax}{argmax}

\begin{document}

\ \vspace{20mm}\\

{\LARGE An Infinite Restricted Boltzmann Machine}

\ \\
{\bf \large Marc-Alexandre C\^ot\'e$^{\displaystyle 1}$, Hugo Larochelle$^{\displaystyle 1}$}\\
{$^{\displaystyle 1}$Department of Computer Science, Universit\'e de Sherbrooke, Sherbrooke, QC, Canada, J1K 2R1}\\


\thispagestyle{empty}
\markboth{}{An Infinite Restricted Boltzmann Machine}
\ \vspace{-0mm}\\
%
\begin{center} {\bf Abstract} \end{center}
We present a mathematical construction for the restricted Boltzmann machine (RBM) that doesn't require specifying the number of hidden units. In fact, the hidden layer size is adaptive and can grow during training. This is obtained by first extending the RBM to be sensitive to the ordering of its hidden units. Then, thanks to a carefully chosen definition of the energy function, we show that the limit of infinitely many hidden units is well defined. As with RBM, approximate maximum likelihood training can be performed, resulting in an algorithm that naturally and adaptively adds trained hidden units during learning. We empirically study the behaviour of this infinite RBM, showing that its performance is competitive to that of the RBM, while not requiring the tuning of a hidden layer size.

\section{Introduction}
  \label{sect:intro}

  Over the years, machine learning research has produced a large variety of latent variable probabilistic models. These include mixture models, factor analysis models, latent dynamical models, and many others. Such models usually require that the dimensionality of the latent representation be specified and fixed during learning. Adapting this quantity is then considered as a separate process, that takes the form of model selection and is normally treated as an additional hyper-parameter to tune.

  For this reason, more recently, there has been a lot of work on extending these models such that the size of the representation can be treated as an adaptive quantity during training. These extensions, often referred to as "infinite" models, are non-parametric in nature where the latent space is infinite with probability 1 and can arbitrarily adapt their capacity to the training data (see \citet{Orbanz2010} for a brief overview).

  While most latent variable models have been extended to one or more infinite variants, a notable exception is the restricted Boltzmann machine (RBM). The RBM is an undirected graphical model for binary vector observations, where the latent representation is itself a binary vector (\ie hidden layer). The RBM (and its extensions to non-binary vectors) have been successfully applied to a variety of problems and data, such as images~\citep{RanzatoM2010}, movie user preferences~\citep{SalakhutdinovR2007}, motion capture~\citep{TaylorG2011}, text~\citep{DahlG2012} and many others. One explanation for the lack of literature on RBMs with an adaptive hidden layer size comes from its undirected nature. Indeed, undirected models tend to be less amenable to a Bayesian treatment of learning, on which relies the majority of the literature on infinite models.

  Our main contribution in this paper is thus a proposal for an infinite RBM which can adapt the effective number of hidden units during training. While our proposal is not based on a Bayesian formulation, it does correspond to the infinite limit of a finite-sized model and behaves in such a way that it effectively adapts its capacity as training progresses.

  First, we propose a finite extension of the RBM that is sensitive to the position of each unit in its hidden layer. This is achieved by introducing a random variable that represents the number of hidden units intervening in the RBM's energy function. Then, thanks to the introduction of an energy cost for using each additional unit, we show that taking the infinite limit of the total number of hidden units is well defined. We describe an approximate maximum likelihood training algorithm for this infinite RBM, based on (Persistent) Contrastive Divergence, which results in a procedure where hidden units are implicitly added as training progresses. Finally, we empirically report how this model behaves in practice and show that it can achieve performance that is competitive to a traditional RBM on the binarized MNIST and Caltech101 Silhouettes datasets, while not requiring the tuning of a hyper-parameter for its hidden layer size.

\section{Restricted Boltzmann Machine}
  \label{sect:rbm}
  \begin{figure}
    \centering
    \includegraphics[width=0.61\textwidth]{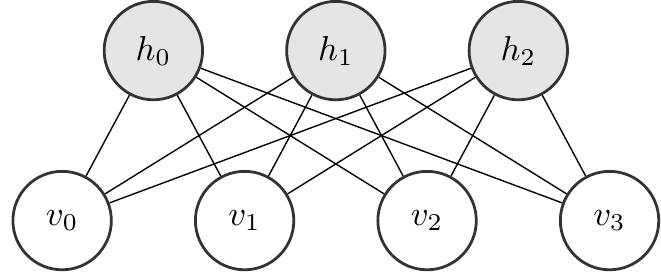}
    \caption[RBM]{
      Graphical model of the restricted Boltzmann Machine. Inter-connections between visible units and hidden units using symmetric weights.
    }
    \label{fig:rbm}
  \end{figure}

  We describe the basic RBM model, which we'll build on to derive its ordered and infinite versions.

  An RBM is a generative stochastic neural network composed of two layers: visible~$\bv$ and hidden~$\bh$. These layers are fully connected to each other, while connections within a layer are not allowed. This means each unit $v_i$ is connected to all $h_j$ units via undirected weighted connections (Figure~\ref{fig:rbm}).

  Given a binary RBM with $D$ visible units and $K$ hidden units, the set of visible vectors is $\cV=\{0,1\}^D$, whereas the set of hidden vectors is $\cH=\{0,1\}^K$. In an RBM model, each configuration $(\bv,\bh)\in\cV\times\cH$ has an associated energy value defined by the following function:
  \begin{equation}
    \label{rbm:energy}
    \Erbm(\bv, \bh) = -\bh^T\bW\bv - \bv^T\bbvis - \bh^T\bbhid
  \end{equation}
  The parameters $\Theta=\{\bW, \bbvis, \bbhid\}$ of this model are the weights $\bW$ ($K\times D$ matrix), the visible unit biases $\bbvis$ ($D\times 1$ vector) and the hidden unit biases $\bbhid$ ($K\times 1$ vector).

  A probability distribution over visible and hidden vectors is defined in terms of this energy function:
  \begin{equation}
    \label{rbm:p_vh}
    \Prbm(\bv, \bh) = \frac{1}{Z} e^{-\Erbm(\bv, \bh)}
  \end{equation}
  with
  \begin{equation}
    \label{rbm:Z}
    Z = \sum_{\bv'\in\cV}\sum_{\bh'\in\cH} e^{-\Erbm(\bv', \bh')}.
  \end{equation}

  We see from Equation~\eqref{rbm:Z} that the partition function $Z$ (normalizing constant) is intractable, as it requires summing over all possible $2^{(D+K)}$ configurations.

  The probability distribution of a visible vector is obtained by marginalizing over all configurations of hidden vectors. One property of the RBM is that the numerator of the marginal $\Prbm(\bv)$ is tractable:
  \begin{equation}
    \label{rbm:p_v}
    \Prbm(\bv) = \frac{1}{Z} \sum_{\bh'\in\cH} e^{-\Erbm(\bv, \bh')} = \frac{1}{Z} e^{-F(\bv)}
  \end{equation}
  with
  \begin{equation}
    \label{rbm:free_energy}
    \Frbm(\bv) = -\bv^T\bbvis -\sum_{i=1}^K \softplus(\bW_{i\cdot}\bv + \bhid_i)
  \end{equation}
  where $\softplus(x) = \ln(1+e^x)$ and the notation $\bW_{i\cdot}$ designates the $i^\text{th}$ row of $\bW$, likewise for columns $\bW_{\cdot j}$. This allows for an equivalent definition of the RBM model in terms of what is known as the free energy $\Frbm(\bv)$. However, the partition function still requires summing over all configurations of visible vectors, which is intractable even for moderate values of $D$.

  RBMs can be learned as generative models, to assign high probability (\ie low energy) to training observations and low probability otherwise. One approach is to minimize the average negative log-likelihood (NLL) for a set of examples $\cD=\{\bv_n\}_{n=1}^N$:
  \begin{equation}
    \label{rbm:loss}
    f(\Theta, \cD) = \frac{1}{N} \sum_{n=1}^N -\ln \Prbm(\bv_n).
  \end{equation}
  The gradient of this objective has a simple form, which is often referred to as the combination of positive and negative phases:
  \begin{equation}
    \label{rbm:gradient}
    \nabla_\theta f(\Theta, \cD) = \underbrace{ \frac{1}{N} \sum_{n=1}^N \nabla_\theta \Frbm(\bv_n) }_\text{Positive phase} - \underbrace{ \sum_{\bv'\in\cV} \Prbm(\bv') \nabla_\theta \Frbm(\bv') }_\text{Negative phase}
  \end{equation}
  where
  \begin{align}
    \label{rbm:grad_W}
    \nabla_\bW \Forbm(\bv) &= -\EE[\bh|\bv]\bv^T = -\widehat{\bh}(\bv) \bv^T\\
    \label{rbm:grad_b}
    \nabla_\bbhid \Forbm(\bv) &= -\EE[\bh|\bv] = -\widehat{\bh}(\bv)\\
    \label{rbm:grad_c}
    \nabla_\bbvis \Forbm(\bv) &= -\bv
  \end{align}
  and where $\widehat{\bh}(\bv) = \bsigm(\bW\bv + \bbhid)$ with $\bsigm(\cdot)$ being the sigmoid function $\sigm(x)=\frac{1}{1+e^{-x}}$ applied element-wise. Derivation for the partial derivatives can be found in Appendix~\ref{AppendixA}.

  Intuitively, the positive phase pushes up the probability of examples coming from our training set, whereas the negative phase lowers the probability of examples generated by the model. Much like the partition function, the negative phase is intractable. To overcome this we approximate the expectation under $\Prbm(\bv)$ with an average of $S$ samples $\cS=\{\hat\bv_s\}_{s=1}^S$ drawn from $\Prbm(\bv)$ \ie the model.
  \begin{equation}
    \label{rbm:gradient_approx}
    \nabla_\theta f(\Theta, \cD) \approx \underbrace{ \frac{1}{N} \sum_{n=1}^N \nabla_\theta \Frbm(\bv_n) }_\text{Positive phase} - \underbrace{ \frac{1}{S} \sum_{s=1}^S \nabla \Frbm(\hat\bv_s)  }_\text{Negative phase}
  \end{equation}
  Moreover, mini-batch training is usually employed and consists in replacing the positive phase average by one over a small subset of the training set, different for every training update.

  Sampling from $\Prbm(\bv)$ can be achieved using block Gibbs sampling, by alternating between sampling $\bv\sim\Prbm(\bv|\bh)$ and $\bh\sim\Prbm(\bh|\bv)$. It can be done efficiently because RBMs have no connections within a layer, meaning that hidden units are conditionally independent given the visible units and vice versa. The conditional distributions of a binary RBM are Bernoulli distributions with parameters
  \begin{align}
    \label{rbm:ph1_v}
    \Prbm(h_i=1|\bv) &= \sigm(\bW_{i\cdot}\bv + \bhid_i)\\
    \label{rbm:pv1_h}
    \Prbm(v_j=1|\bh) &= \sigm(\bh^T\bW_{\cdot j} + \bvis_j)
  \end{align}

  In theory, the Markov chain should be run until equilibrium before drawing a sample for every training update, which is highly inefficient. Thus, Contrastive Divergence (CD) learning is often employed, where we initialize the update's Gibbs chains to the training examples and only perform $T$ steps of Gibbs sampling~\citep{Hinton2002}. Another approach, referred to as stochastic approximation or Persistent CD (PCD)~\citep{Tieleman2008}, is to not reinitialize the Gibbs chains between updates.

\section{Ordered Restricted Boltzmann Machine}
  \label{sect:orbm}
  \begin{figure}
    \centering
    \includegraphics[width=0.61\textwidth]{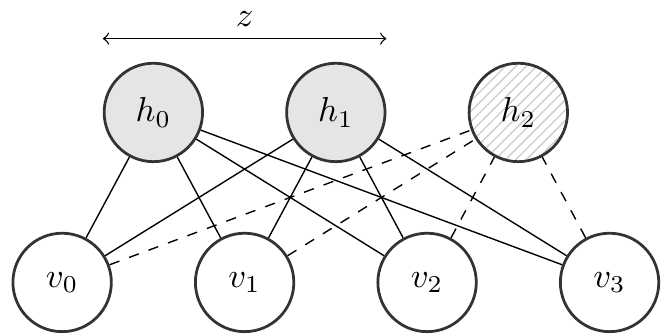}
    \caption[oRBM]{
      Illustration of the ordered RBM. Since $z=2$ only the first two hidden units are selected.
    }
    \label{fig:orbm}
  \end{figure}

  The model we propose is a variant of the RBM where the hidden units $\bh$ are ordered from left to right, with this order being taken into account by the energy function. We refer to this model as an ordered RBM (oRBM). As shown in Figure~\ref{fig:orbm}, the oRBM takes hidden unit order into account by introducing a random variable $z$ that can be understood as the effective number of hidden units participating to the energy. Hidden units are selected starting from the left and the selection of each hidden unit is associated with an incremental cost in energy.

  Concretely, we define the energy function of the oRBM as
  \begin{equation}
    \label{orbm:energy}
    \Eorbm(\bv,\bh,z) = -\bv^T\bbvis -\sum_{i=1}^{z} \left( h_i(\bW_{i\cdot}\bv + \bhid_i) - \beta_i \right)
  \end{equation}
  where $z$ represents the number of selected hidden units that are active and $\beta_i$ is a energy penalty for selecting each $i^\text{th}$ hidden unit. As we will see, carefully parametrizing the per unit energy penalty will allow us to consider the case of an infinite pool of hidden units.

  In our experiments, as we wanted the filters of each unit to be the dominating factor in a unit being selected, we parametrized it as $\beta_i=\beta\softplus(\bhid_i)$, where $\beta$ is a global hyper-parameter (critically, as we'll discuss later, this hyper-parameter doesn't actually require tuning and a generic value for it works fine). Intuitively, the penalty term acts as a form of regularization since it forces the model to avoid using more hidden units than needed, prioritizing smaller networks.

  Moreover, having the penalty depending on the hidden biases also implies that the selection of a hidden units (\ie influencing the outcome of the random variable $z$) will be mostly controlled by the values taken by the connections $\bW$.
  Higher values of the bias of a hidden unit will not increase its probability of being selected. In other words, for the model to increase its capacity and better fit the training data, it will have to learn better filters. Note that alternative parametrizations could certainly be considered.

  As with the RBM,
  $\Porbm(\bv)$ is defined in terms of its energy function. For this, we have to specify the set of legal values for $\bv$, $\bh$ and $z$. Since, for a given $z$, the value of the energy is irrelevant for the dimensions of $\bh$ from $z$ to $K$, we will assume they are set to 0. There is thus a coupling between the value of $z$ and the legal values of $\bh$. We will note ${\cH_z = \{\bh \in \cH | h_k = 0~\forall k>z\}}$ the legal values of $\bh$ for a given $z$. As for $z$, it can vary in $\{1,\ldots,K\}$, and $\bv\in \cV$ as usual.

  The joint probability over $\bv$, $\bh$ and $z$ is thus:
  \begin{equation}
    \label{orbm:p_vhz}
    \Porbm(\bv, \bh, z) = \frac{1}{Z} e^{-\Eorbm(\bv, \bh, z)}
  \end{equation}
  where
  \begin{equation}
    \label{orbm:Z}
    Z = \sum_{z'=1}^K \sum_{\bv'\in\cV} \sum_{\bh'\in\cH_{z'}} e^{-\Eorbm(\bv', \bh', z')}.
  \end{equation}
  As for the marginal distribution $\Porbm(\bv)$ of the oRBM model, it can also be written in terms of a free energy. Indeed, in a derivation similar to the case of the RBM, we can show:
  \begin{align}
    \label{orbm:p_v_z_free_energy}
    \Porbm(\bv) &= \frac{1}{Z} \sum_{z=1}^K \sum_{\bh\in\cH_z} e^{-\Eorbm(\bv, \bh, z)}  = \frac{1}{Z} \sum_{z=1}^K e^{-F(\bv,z)} \\
    \label{orbm:v_z_free_energy}
    \Forbm(\bv,z) &= -\bv^T\bbvis -\sum_{i=1}^z \left( \softplus(\bW_{i\cdot}\bv + \bhid_i) - \beta_i \right)
    \end{align}
  This gives us a free energy where only the hidden units have been marginalized. We can also derive a formulation where the free energy depends only on $\bv$:
  \begin{equation}
    \label{orbm:p_v_free_energy}
    \Porbm(\bv) = \frac{1}{Z} \sum_{z=1}^K e^{-F(\bv,z)} = \frac{1}{Z} e^{-F(\bv)}
    \quad\text{with}\quad
    \Forbm(\bv) = -\ln\left(\sum_{z=1}^K e^{-F(\bv,z)}\right)
  \end{equation}
  It should be noticed that, in the oRBM, $z$ does not correspond to the number of hidden units assumed to have generated {\it all} observations. Instead, the model allows for different observations having been generated by a different number of hidden units. Specifically, for a given $\bv$, the conditional distribution over the corresponding value of $z$ is
  \begin{equation}
  \label{orbm:p_z_given_v}
    \Porbm(z|\bv) = \frac{\exp(-\Forbm(\bv, z))}{\sum_{z'=1}^K \exp(-\Forbm(\bv, z'))}~.
  \end{equation}
  As for the conditional distribution over the hidden units, given a value of $z$ it takes the same form as for the regular RBM, except for unselected hidden units which are forced to zero. Similarly, the distribution of $\bv$ given a value of the hidden layer and $z$ reflects that of the RBM:
  \begin{align}
    \label{orbm:conditionals}
    \Porbm(h_i=1|\bv,z) &=
    \begin{aligned}
      \begin{cases}
        \sigm(\bW_{i\cdot}\bv + \bhid_i) & \text{if $i\le z$}\\
        0 & \text{otherwise}
      \end{cases}
    \end{aligned}
    \\
    \Porbm(v_j=1|\bh,z) &= \sigm\left(\sum_{i=1}^{z} W_{ij}h_i + \bvis_j\right)
  \end{align}

  To train the oRBM, we can also rely on CD or PCD for estimating the gradients based on Equation~\ref{rbm:gradient_approx} but using $\Forbm(\bv)$ as defined in equation~\ref{orbm:p_v_free_energy}.
  Defining $\ind_z = [\overbrace{1, \dots, 1}^z, 0, \dots, 0]^T$ and
  $\mathbf{cdf}(z|\bv) = [\Porbm(z<1|\bv),\dots,\Porbm(z<K|\bv)]^T$
  with $\odot$ denoting the element-wise product,
  the free energy gradients are then slightly modified as follows:

  \begin{align}
    \label{orbm:grad_W}
    \nabla_\bW \Forbm(\bv)
      &= -\EE_{\bh,z}[\bh\odot \ind_{z}|\bv]\bv^T
      = -(\widehat{\bh}(\bv) \odot (1-\mathbf{cdf}(z|\bv)))\bv^T \\
    \label{orbm:grad_b}
    \nabla_\bbhid \Forbm(\bv) &= -\EE_{\bh,z}[(\bh - \beta\bsigm(\bbhid))\odot \ind_{z}|\bv]
    = -(\widehat{\bh}(\bv) - \beta\bsigm(\bbhid)) \odot (1-\mathbf{cdf}(z|\bv)) \\
    \label{orbm:grad_c}
    \nabla_\bbvis \Forbm(\bv) &= -\bv
  \end{align}
  with $\widehat{\bh}(\bv) = \bsigm(\bW\bv + \bbhid)$. Derivation for the partial derivatives can be found in Appendix~\ref{AppendixA}.

  Compared to the RBM, computing these gradients requires one additional quantity: the vector of cumulative probabilities $\mathbf{cdf}(z|\bv)$. Fortunately, this quantity can be efficiently computed, in $\BigO(K)$, by first computing the required probabilities vector $\Porbm(z|\bv)$ and performing a cumulative sum.

  Sampling from $\Porbm(\bv)$ slightly differs from the RBM as we need to consider $z$ in the Markov chain. With the oRBM, Gibbs steps alternate between sampling $(\bh,z)\sim\Porbm(\bh,z|\bv)$ and $\bv\sim\Porbm(\bv|\bh,z)$. Sampling from $\Porbm(\bh,z|\bv)$ is done in two steps: $z\sim\Porbm(z|\bv)$ followed by $\bh\sim\Porbm(\bh|\bv,z)$.

  During training, what we observe is that the hidden units are each trained gradually, in sequence, from left to right. This effect is mainly due to the multiplicative term ${(1-\mathbf{cdf}(z|\bv))}$ in the hidden unit parameter updates of Equations~\ref{orbm:grad_W}~and~\ref{orbm:grad_b}, which is monotonically decreasing. Effectively, the model is thus growing in capacity during training, until its maximum capacity of $K$ hidden units.

\section{Infinite Restricted Boltzmann Machine}
  \label{sect:irbm}

  \begin{figure}
    \centering
    \includegraphics[width=0.61\textwidth]{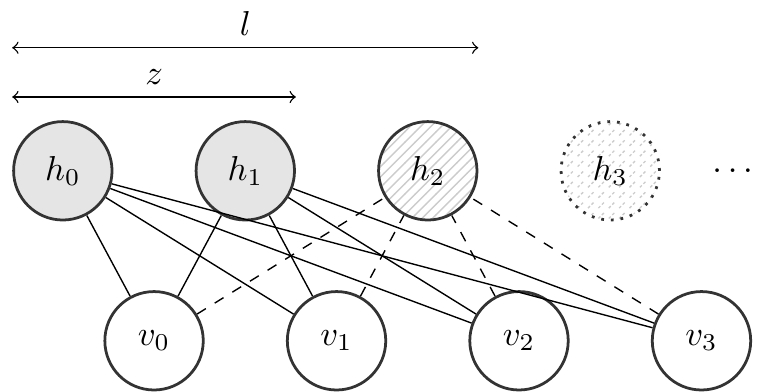}
    \caption[iRBM]{
      Illustration of the infinite RBM. With $z=2$, only the first two hidden units are currently selected. The dashed lines illustrate that there are connections that are trained (non-zero) with the third hidden unit. All (infinitely many) hidden units after the third have zero-valued weights, which correspond to $l$ being equal to 3.
    }
    \label{fig:irbm}
  \end{figure}

  The growing behaviour of the oRBM begs for the question: could we achieve a similar effect without having to specify a maximum capacity to the model? Indeed, while \citet{Montufar2011} have shown that with $2^{V-1}-1$ hidden units an RBM is a universal approximator, a variant of the RBM that could automatically increase its capacity until it is sufficiently high is likely to yield much smaller models in practice.

  It turns out that this is possible, by taking the limit of $K\rightarrow \infty$. For this reason, we refer to this model as the infinite RBM (iRBM).

  This limit is made possible thanks to two modeling choices. The first is the assumption that a finite (but variable!) number of hidden units have non-zero weights and biases. This is trivial to ensure, for any optimization procedure, using any amount of any type of weight decay (\eg L2 or L1 regularization) on all the weights and hidden biases. An infinite number of non-zero weights and biases could then correspond to an infinite penalty, so no proper optimization would ever diverge to this solution, no matter the initialization. This is guaranteed when using L1 regularization, thanks to its sparsity inducing property. As for L2 regularization, while it could theoretically lead to an infinite number of hidden units (\eg if the L2 norm of the parameters associated with each hidden unit decreases exponentially with respect to the position of the hidden unit), in practice the floating precision would clip very small parameters to zero, thus having a finite number of hidden units.

  The second key choice is our parametrization of the per-unit energy penalty $\beta_i$, which will ensure that the infinite sums required in computing probabilities will be convergent. For instance, consider the conditional $\Pirbm(z|\bv)$:
  \begin{align}
  \label{irbm:p_z_given_v}
    \Pirbm(z|\bv) = \frac{\exp(-\Firbm(\bv, z))}
                           { Z(\bv) }
                    = \frac{\exp(-\Firbm(\bv, z))}
                           { \sum_{z'=1}^\infty \exp(-\Firbm(\bv, z')) }
  \end{align}
  Let's note $l$ the number of effectively trained hidden units, \ie where all hidden units $>l$ have zero weights and biases. This is guaranteed to happen thanks to the growing behaviour that ensures hidden units are ordered from left to right. Then, we can split the normalization constant  $Z(\bv)$ of Equation~\ref{irbm:p_z_given_v} into two parts, split at $z=l$, as follows:
  \begin{align}
    Z(\bv) &=\sum_{z=1}^l \exp( -\Firbm(\bv, z) ) + \sum_{z=l+1}^\infty \exp( -\Firbm(\bv, z) ) \nonumber\\
           &= \sum_{z=1}^l \exp( -\Firbm(\bv, z) )
           + \sum_{z=l+1}^\infty \exp\left( -\Firbm(\bv, l) + \sum_{i=l+1}^z \softplus(\bW_{i\cdot}\bv + \bhid_i) - \beta_i\right) \nonumber\\
           &= \sum_{z=1}^l \exp( -\Firbm(\bv, z) )
           + \exp( -\Firbm(\bv, l) ) \underbrace{\sum_{z=1}^\infty \exp((1-\beta)\softplus(0))^z}_\text{Geometric series}\label{irbm:geom}
  \end{align}
  where Equation~\ref{irbm:geom} is obtained by exploiting the fact that all weights and biases of hidden units at position $l+1$ and higher are zero. By ensuring that $\beta>1$, the geometric series of Equation~\ref{irbm:geom} is finite and can be analytically computed. This in turn implies that $\Pirbm(z|\bv)$ is tractable and can be sampled from.
  Following a similar reasoning, the global partition function $Z$ can be shown to be finite (see Appendix~\ref{AppendixB}), thus yielding a properly defined joint distribution for any configurations with a finite number of non-zero weights and hidden biases.

  One could think that, compared to a regular RBM, we have merely traded the hyper-parameter of the hidden layer size with the hyper-parameter $\beta$. However, crucially, $\beta$'s role is only to ensure that the iRBM is properly defined, and the penalty it imposes in the energy function can be compensated by the learned parameters. The extent to which the parameters can grow enough to compensate for that penalty is then controlled by the strength of weight decay, a hyper-parameter the iRBM shares with the RBM. We've thus effectively removed one hyper-parameter.
  Moreover, we've indeed observed that results are robust to the choice of $\beta$, that is finely tuning beta was not necessary to ultimately achieving good performance. While the choice of $\beta$ can impact the number of epochs it would take for the weights to compensate for the penalty, this (the number of epochs) is a quantity that must be tuned anyways, even in regular RBMs.

  The question of the identifiability of the binary RBM is a complex one, which has been studied~\citep{Cueto2010}. Unlike the RBM, the iRBM is sensitive to the ordering of its hidden units, thanks to the penalty term. This means permutations of iRBM's hidden units do not correspond to the same distribution, making its parametrization more identifiable.

  As for learning, it can be done mostly by following the procedure of the oRBM, \ie minimizing the NLL with stochastic gradient descent using (Persistent) CD to approximate the gradients.
  One slight modification is required however. Indeed, since the free energy gradient for the hidden weights and biases can be non-zero for all (infinite) hidden units, we cannot use the gradient of Equations~\ref{orbm:grad_W}~and~\ref{orbm:grad_b} for all hidden units.

  To avoid this issue, we consider the following observation. Instead of using the derivative of $\Frbm(\bv)$, we could instead use the derivative of $\Frbm(\bv,z)$, where $z$ is obtained by sampling from $\Porbm(z|\bv)$:
  \begin{align}
    \label{orbm:grad_W_inf}
    \nabla_\bW \Forbm(\bv,z) &= -\EE_{\bh}[\bh\odot \ind_{z}|z,\bv]\bv^T
     = -(\widehat{\bh}(\bv) \odot \ind_{z})\bv^T\\
    \label{orbm:grad_b_inf}
    \nabla_\bbhid \Forbm(\bv,z) &= -\EE_{\bh}[(\bh - \beta\bsigm(\bbhid))\odot \ind_{z}|z,\bv]
    = -(\widehat{\bh}(\bv) - \beta\bsigm(\bbhid)) \odot \ind_{z} ~.
  \end{align}

  In this case, all weights and biases with an index greater than the sampled $z$ have a gradient of zero, \ie do not require any update. Moreover, the expectation of these gradients with respect to $z$ (conditioned on $\bv$) are the gradients of $\Frbm(\bv)$, making them unbiased in this respect. This comes at the cost of higher variance in the updates. But thanks to this observation, we are justified to use a hybrid approach, where we use the $\Frbm(\bv)$ gradients only for the units with index less or equal than $l$, and "use" the gradient of $\Frbm(\bv,z)$ for the other units, \ie leave them set to zero.

  As previously mentioned, we use weight decay to ensure that the number of non-zero parameters cannot diverge to infinity.
  For practical reasons, our implementation also used a capacity-limiting heuristic. If the Gibbs sampling chain ever sampled a value for $z$ that is greater than $l$, then we clamped it to $l+1$. Intuitively, this corresponds to "adding" a single hidden unit. This avoids filling all the memory in the (unlikely) event where we'd draw a large value for $z$. When adding a hidden unit, its associated weights and biases are initialized to zero.

  We emphasize that these were not required to avoid divergence (weight decay is sufficient): it merely ensured a practical and efficient implementation of the model on the GPU.
  Note also that when using L1 regularisation, $l$ can decrease in value, thanks to the sparsity promoting property of the L1 norm.
  Again, we highlight that while a finite number of weights and biases is maintained, that number of such weights does vary and is learned, while the implicit number of hidden units is indeed infinite (infinitely many contribute to the partition function).

\section{Related Work}
  This work falls within the research literature on discovering extensions of the original RBM model to different contexts and objectives. Of note here is the implicit mixture of RBMs~\citep{Nair2008}.
  Indeed, the oRBM can be interpreted as a special case of an implicit mixture of RBMs. Writing $\Porbm(\bv)$ as $\sum_{z=1}^K \Porbm(z) \Porbm(\bv|z)$
  we see that the oRBM is an implicit mixture of $K$ RBMs, where each RBM has a different number of hidden units (from 1 to $K$) and the weights are tied between RBMs. The prior $\Porbm(z)$ represents the probability of using the $z^\text{th}$ RBM and is also derived from the energy function. However, as in the implicit mixture of RBMs, $\Porbm(z)$ is intractable as it would require the value of the partition function. That said, the work of~\citet{Nair2008} is otherwise very different and did not address the question of having an RBM with adaptive capacity.

  Another related work is that of the Cardinality RBMs proposed by~\citet{Swersky2012}. They used a cardinality potential to control the sparsity of the RBM, \ie limiting the number of hidden units that can be active.
  In the oRBM and the iRBM, $z$ effectively acts as an upper bound on the number of hidden units $h_i$ that can be equal to 1, since we are limiting $\bh$ to be in $\cH_z$, a subset of $\cH$. In their work, \citet{Swersky2012} use cardinality potentials that allow only configurations having at most $k$ active hidden units. One difference with our work however is that their cardinality potential is order agnostic, meaning that the active hidden units can be positioned anywhere within the hidden layer while still satisfying the cardinality potential. On the other hand, in the oRBM, all units with index higher than $z$ must be set to zero, with only the previous hidden units being allowed to be active. In addition, their parameter $k$ is fixed during training whereas our number of active hidden units $z$ changes depending on the input.

  The oRBM also bears some similarity with autoencoders trained by a nested version of dropout~\citep{Rippel2014}. Nested dropout works by stochastically selecting the number of hidden units used to reconstruct an input example at training time, and so independently for each update and example. \citet{Rippel2014} showed that this defines a learning objective that makes the solution identifiable and no longer invariant to hidden unit permutation. In addition to being concerned with a different type of model, this work doesn't discuss the case of an unbounded and adaptive hidden layer size.

  \citet{WellingM2003} proposed a self supervised boosting approach, which is applicable to the RBM and in which hidden units are sequentially added and trained. However, like boosting in general and unlike the iRBM, this procedure trains each hidden unit greedily instead of jointly, which could lead to much larger networks than necessary. Moreover, it is not easily generalizable to online learning.

  While the work on unsupervised neural networks with adaptive hidden layer size is otherwise relatively scarse, there's been much more work in the context of supervised learning. There is the well known work of \citet{FahlmanS1990} on Cascade-Correlation networks. More recently, \citet{ZhouG2012} proposed a procedure for learning discriminative features with a denoising autoencoder (a model related to the RBM). The procedure is also applicable to the online setting. It relies on invoking two heuristics that either add or merge hidden units during training.
  We note that the iRBM framework could easily be generalized to discriminative and hybrid training as in \citet{ZhouG2012}.
  The corresponding mecanisms for adding and merging units would then be implicitly derived from gradient descent on the corresponding supervised training objective.

  Finally, we highlight that our model is not based on a Bayesian formulation, as most of the literature on infinite models. On the other hand, it does correspond to the infinite limit of a finite-sized model and yields a model that can learn its size with training.

\section{Experiments}
  \label{sect:experiments}

  We compare the performance of the oRBM and the iRBM with the classic RBM on two datasets: binarized MNIST~\citep{Salakhutdinov2008} and CalTech101 Silhouettes~\citep{Marlin2010a}. We aim to demonstrate that the iRBM effectively removes the need of tuning an hyper-parameter for the hidden layer size while still achieving comparable performance to the standard RBM.
  The code to reproduce the experiments of the paper is available on GitHub\footnote{\url{http://github.com/MarcCote/iRBM}}. Our implementation is done using Theano~\citep{Bastien2012,Bergstra2010}.

  For completeness, we wish to mention that more sophisticated or deep models have reported results on one or both of these datasets (\eg EoNADE~\citep{Uria2014}, DBNs~\citep{Murray2009}, Deep autoregressive networks~\citep{Gregor2014}, Iterative Neural Autoregressive Distribution Estimator~\citep{Raiko2014}) that improve on the standard RBM. However, since our objective with the iRBM is to effectively remove a hyper-parameter of the RBM, instead of achieving improved performances, we focus our comparison on this baseline.

  All NLL results of this section were obtained by estimating the log-partition function $\ln\hat{Z}$ using Annealed Importance Sampling (AIS)~\citep{Salakhutdinov2008} with 100,000 intermediate distributions and 5000 chains. As an additional validation step, samples were generated from best models and visually inspected.

  Each model was trained with mini-batch stochastic gradient descent using batch size of 64 examples and using PCD with 10 Gibbs steps between parameter updates. We used the ADAGRAD stochastic gradient update~\citep{Duchi2010}, a per-dimension learning rate method, to train the oRBMs and the iRBMs. We found that having different learning rates for different hidden units was very beneficial, since units positioned earlier in the hidden layer will approach convergence faster than units to their right, and thus will benefit from a learning rate decaying more rapidly. We tried several learning rates $lr \in \{5\!\!\times\!\!10^{-1}, 10^{-1}, 5\!\!\times\!\!10^{-2}, 10^{-2}\}$ and always set ADAGRAD's epsilon parameter to $10^{-6}$.

  We also tested different values for both L1 and L2 regularization's factor $\lambda~\in~\{0, 10^{-2}, 10^{-3}, 10^{-4}, 10^{-5}\}$. Note that we allow the iRBM to shrink only if L1 regularization is used.

  We did try varying the $\beta$ found in the penalty term and as expected we've found results to be robust to its value. Since $\beta$ must be greater than 1, we explored positive constants to add to 1, on a log scale (1, 0.25, 0.1, 0.01, 0.001, etc.).
  We settled on using $\beta=1.01$ for all experiments as it provides a penalty high enough to have a growing behavior and requires around five hundred epochs for the weights to compensate for the penalty.

  Finally, we note that improved performances could certainly have been achieved using an improved sampler (\eg parallel tempering~\citep{Desjardins2010}) or parametrization (\eg enhanced gradient parametrization~\citep{Cho2013}). However, these changes would equally improve the baseline RBM, so we decided to concentrate on this more common learning setup.

  \subsection{Binarized MNIST}
    \begin{table}
      \renewcommand{\baselinestretch}{1}
      \caption{Average NLL on binarized MNIST test set for best RBMs, oRBM and iRBM. Partition functions were estimated using AIS with 100,000 intermediate distributions and 5000 chains. The confidence interval on the average NLL assumes $\ln\hat{Z}$ has no variance and reflects the confidence of a finite sample average. By taking the uncertainty about the partition function into account, the interval would be larger.}
      \label{tab:results_mnist}
      \begin{center}
          \begin{sc}
            \begin{tabular}{rcccc}
              \toprule
                & & \multicolumn{3}{c}{Binarized MNIST} \\
                \cmidrule(lr){3-5}
                Model & Size & $\ln\hat{Z}$ & $\ln(\hat{Z}\pm3\sigma)$ & Avg. NLL\\
                \midrule
                RBM & 100 & 600.92 & [600.88, 600.95] & 98.17 $\pm$ 0.52 \\
                RBM & 500 & 613.28 & [613.24, 613.31] & 86.50 $\pm$ 0.44 \\
                RBM & 2000 & 1099.07 & [1098.94, 1099.17] & 85.03 $\pm$ 0.42 \\
                \cmidrule(lr){1-5}
                oRBM & 500 & 40.06 & [39.90, 40.19] & 88.15 $\pm$ 0.46 \\
                iRBM & 1208 & 40.32 & [40.03, 40.54] & 85.65 $\pm$ 0.44 \\
              \bottomrule
            \end{tabular}
          \end{sc}
      \end{center}
    \end{table}

    \begin{figure}
      \centering
      \subfigure[RBM]{
        \includegraphics[width=0.45\textwidth]{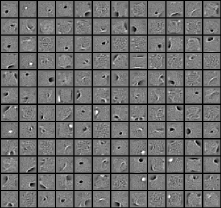}
      }
      \subfigure[iRBM]{
        \includegraphics[width=0.45\textwidth]{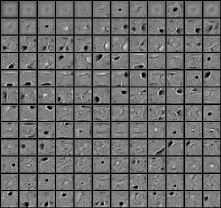}
      }
      \caption[RBM vs. iRBM filters]{
        Comparing the filters of an RBM and an iRBM both trained on binarized MNIST. The first 96 filters are shown starting from the top-left corner and incrementing across columns first.
      }
      \label{fig:mnist_filters}
    \end{figure}

    \begin{figure}
      \centering
      \subfigure[Test set]{
        \includegraphics[width=0.21\columnwidth]{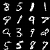}
      }
      \subfigure[RBM]{
        \includegraphics[width=0.21\columnwidth]{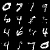}
      }
      \subfigure[oRBM]{
        \includegraphics[width=0.21\columnwidth]{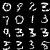}
      }
      \subfigure[iRBM]{
        \includegraphics[width=0.21\columnwidth]{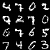}
      }
      \caption[RBM vs. oRBM vs. iRBM samples]{
        Comparison between data from binarized MNIST and random samples generated from the three models by randomly initializing visible units and running 10,000 Gibbs steps. The RBM and oRBM both have 500 hidden units, whereas the iRBM final size is 1208 hidden units.
      }
      \label{fig:mnist_samples}
    \end{figure}

    \begin{figure}
      \centering
      \subfigure[oRBM]{
        \includegraphics[width=0.37\textwidth]{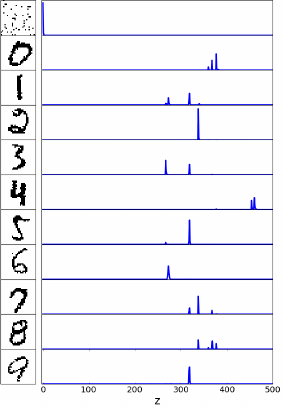}
      }
      \subfigure[iRBM]{
        \includegraphics[width=0.37\textwidth]{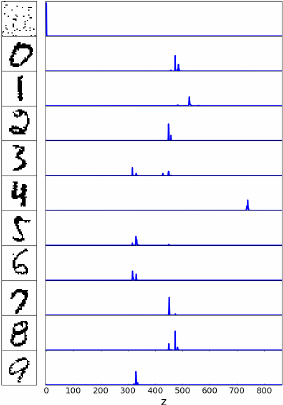}
      }
      \caption[Conditional distribution of $z$ given $\bv$]{
        Each row shows a plot of $\Porbm(z|\bv)$ where $\bv$ is a given example from MNIST test set and is displayed to the left. The first row illustrates the impact of a noisy image on sampling $z$. As explained in Section~3 of the paper, we see that different input images are related to different values for the number $z$ of used units.
      }
      \label{fig:pz_given_v_mnist}
    \end{figure}

    \begin{figure}
      \centering
      \includegraphics[width=0.99\textwidth]{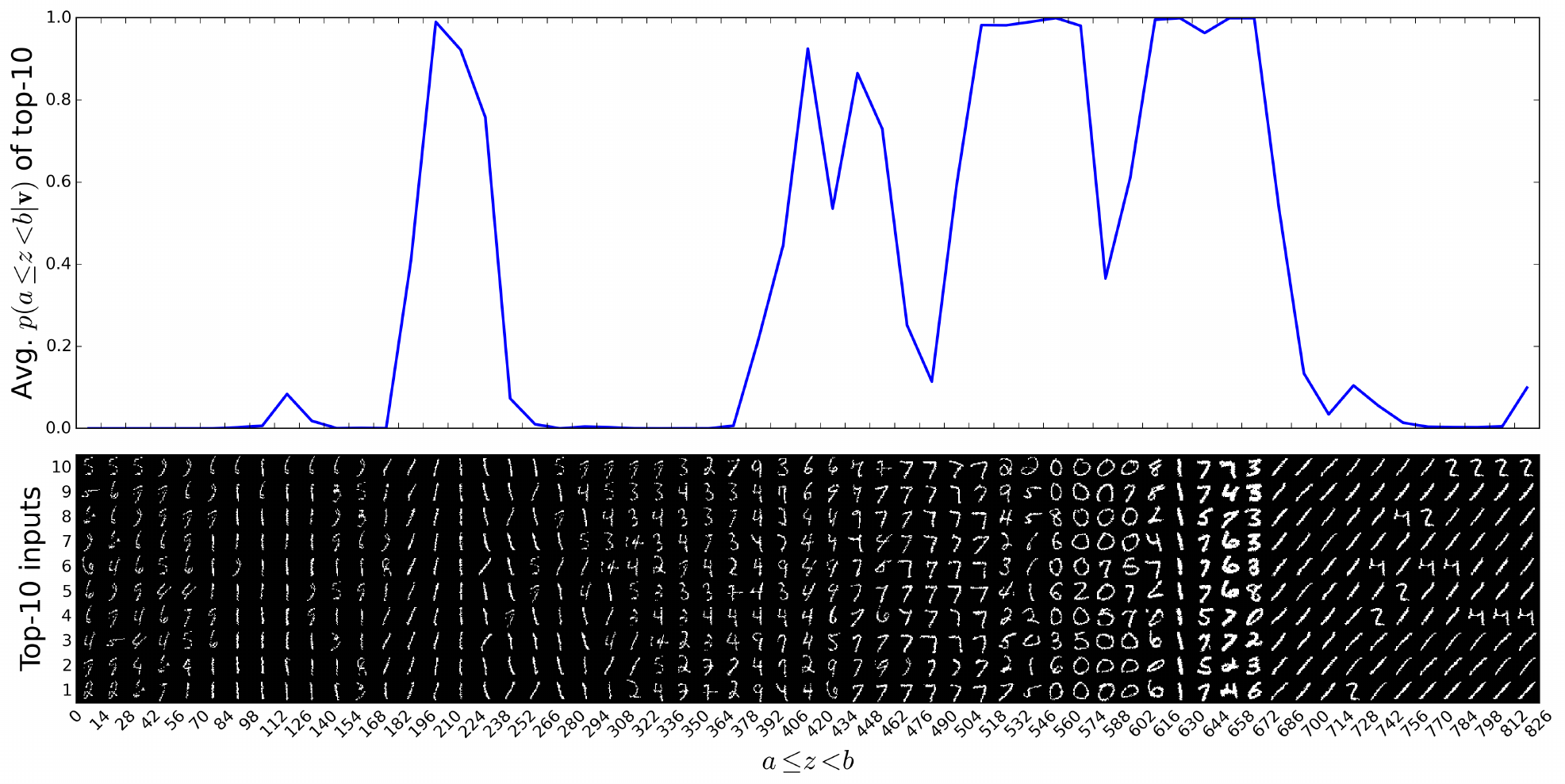}
      \caption[Top10 inputs maximizing different regions of $\Porbm(z|\bv)$]{
        {\bf(Bottom)} Top 10 inputs from the test set with highest value of $\Porbm(z|\bv)$ within different intervals for $z$, \ie $\argmax_{\bv}\Porbm(a\leq z < b|\bv)$ for different intervals $[a,b)$. Interestingly, bolder inputs seem to be related to bigger values for the number $z$ of used units. Also, simpler characters (\eg "ones") tend to favor smaller values of $z$ compared to more complex characters. {\bf(Top)} Average of $\Porbm(a\leq z < b|\bv)$ over the top 10 inputs. Low values highlights regions in the hidden layer where the hidden units are only useful when taken together with hidden units further right in the layer.
      }
      \label{fig:topk_prob_z_given_x}
    \end{figure}

    The MNIST dataset\footnote{\url{http://yann.lecun.com/exdb/mnist}} is composed of 70,000 images of size 28x28 pixels representing handwritten digits (0-9). Images have been stochastically binarized according to their pixel intensity as in~\citet{Salakhutdinov2008}. We use the same split as in \citet{Larochelle2011}, corresponding to 50,000 examples for training, 10,000 for validation and 10,000 for testing.

    Each model was trained up to 5000 epochs but we performed AIS evaluation every 1000 epochs and kept the model having the best NLL approximation on the valid set. We report the associated NLL approximations obtained on the test set. Taking after past studies assessing RBM results on binarized MNIST, we fixed the number of hidden units to 500 for the RBM and the oRBM. Best results for the RBM, oRBM and iRBM are reported in Table~\ref{tab:results_mnist}. The oRBM and the iRBM models reach competitive performance compared to the RBM. Samples from all three models are illustrated in Figure~\ref{fig:mnist_samples}.

    The best RBM (500 hidden units) was trained without any regularization and $lr\!=\!10^{-2}$ for 5000 epochs.
    We used our own implementation to train the RBM, which is why our result slightly differs from what is reported by \citet{Salakhutdinov2008}. The difference can be justified by the fact that they used the full 60,000 training set images, instead of a 50,000 subset. Also, they use a custom schedule to gradually increase the number of CD steps during training. That said, the oRBM and the iRBM would probably also benefit from having more training data and an improved sampling strategy.

    The best oRBM (500 hidden units) was trained without any regularization and $lr\!=\!3\!\!\times\!\!10^{-2}$ for 500 epochs. After 3000 epochs, the best iRBM had 1208 hidden units with non-zero weights. It was trained with L1 regularization using a regularization factor of $\lambda\!=\!10^{-4}$ and $lr\!=\!5\!\!\times\!\!10^{-2}$.

    To show that our best iRBM does find an appropriate number of hidden units, we compared it with two other RBMs having respectively 100 and 2000 hidden units. Both were trained for 5000 epochs without any regularization and respectively with $lr\!=\!10^{-1}$ and $lr\!=\!10^{-2}$. Results are reported in Table~\ref{tab:results_mnist} where we can see the oRBM and the iRBM still achieve competitive results compared to the RBM with 2000 hidden units.

    Figure~\ref{fig:mnist_filters} shows the ordering effect on the filters obtained with an iRBM. The ordering is even more apparent when observing the hidden unit filters during training. We generated a video of this visualization, illustrating the filter values and the generated negative samples at epochs 1, 10, 50 and 100. See link: \url{http://goo.gl/LGQDaI}.

    Interestingly, we've observed that Gibbs sampling can mix much more slowly with the oRBM. The reason is the addition of variable $z$ increases the dependence between states and thus hurts the convergence of Gibbs sampling. In particular, we observed that when the Gibbs chain is in a state corresponding to a noisy image without any structure, it can require many steps before stepping out of this region of the input space. Yet, comparing the free energy of such random images and images that resemble digits confirmed that these random images have significantly higher free energy (and thus are unlikely samples of the model). Figure~\ref{fig:pz_given_v_mnist} also confirms the high dependence between $z$ and $\bv$: the distribution of the unstructured image is peaked at $z=1$, while all digits prefer values of $z$ greater than 250. To fix this issue, we've found that simply initializing the Gibbs chain to $z=K$ was sufficient. We used this when sampling from a trained oRBM model.

    The iRBM doesn't seem to suffer as much from a low mixing rate and thus doesn't require the $z=K$ initialization heuristic for sampling. In fact, using the heuristic when sampling from an iRBM has almost no impact on the final samples when running 10,000 Gibbs steps. This could be an artefact of the model being trained progressively, \ie we only add one hidden unit when sampling a large value for $z$ bigger than $l$. Understanding how the lower mixing rate affects the proposed models and if a heuristic such as the one we mentioned earlier could be used to improve training is a topic left for future work.

    We've also investigated what kind of inputs are maximizing $\Porbm(z|\bv)$, for different values of $z$. Using our best iRBM model trained with L1 regularization, we generated Figure~\ref{fig:topk_prob_z_given_x}. It highlights the fact that $\Porbm(z|\bv)$ does capture some structure about the data, as the identity of the character with highest $\Porbm(z|\bv)$ vary between different values of $z$.

  \subsection{CalTech101 Silhouettes}

    \begin{table}
      \renewcommand{\baselinestretch}{1}
      \caption{Average NLL on CalTech101 Silhouettes test set estimated using AIS with 100,000 intermediate distributions and 5000 chains. The confidence interval on the average NLL assumes $\ln\hat{Z}$ has no variance and reflects the confidence of a finite sample average. By taking the uncertainty about the partition function into account, the interval would be larger.}
      \label{tab:results_caltech}
      \begin{center}
          \begin{sc}
            \begin{tabular}{rcccc}
              \toprule
                & & \multicolumn{3}{c}{CalTech101 Silhouettes} \\
                \cmidrule(lr){3-5}
                Model & Size & $\ln\hat{Z}$ & $\ln(\hat{Z}\pm3\sigma)$ & Avg. NLL\\
                \midrule
                RBM & 100 & 2512.20 & [2511.62, 2512.56] & 177.37 $\pm$ 2.81 \\
                RBM & 500 & 2385.91 & [2385.68, 2386.10] & 119.05 $\pm$ 2.27 \\
                RBM & 2000 & 3353.47 & [3349.85, 3354.15] & 118.29 $\pm$ 2.25 \\
                \cmidrule(lr){1-5}
                oRBM & 500 & 1782.96 & [1782.88 1783.02] & 114.99 $\pm$ 1.97 \\
                iRBM & 915 & 2000.08 & [1999.93, 2000.22] & 121.47 $\pm$ 2.07 \\
              \bottomrule
            \end{tabular}
          \end{sc}
      \end{center}
    \end{table}

    \begin{figure}
      \centering
      \subfigure[Test set]{
        \includegraphics[width=0.21\columnwidth]{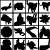}
      }
      \subfigure[RBM]{
        \includegraphics[width=0.21\columnwidth]{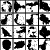}
      }
      \subfigure[oRBM]{
        \includegraphics[width=0.21\columnwidth]{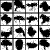}
      }
      \subfigure[iRBM]{
        \includegraphics[width=0.21\columnwidth]{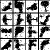}
      }
      \caption[RBM vs. oRBM vs. iRBM samples]{
        Comparison between data from CalTech101 Silhouettes and random samples generated from three models by randomly initializing visible units and running 10,000 Gibbs steps. The RBM and oRBM both have 500 hidden units, whereas the iRBM final size is 915 hidden units.
      }
      \label{fig:caltech_samples}
    \end{figure}

    The CalTech101 Silhouettes dataset\footnote{\url{http://people.cs.umass.edu/~marlin/data.shtml}}~\citep{Marlin2010a} is composed of 8,671 images of size 28x28 binary pixels, representing object silhouettes (101 classes). The dataset is divided in three subsets: 4,100 examples for training, 2,264 for validation and 2,307 for testing.

    Following a protocol similar to the one used for MNIST, each model was trained up to 5000 epochs, AIS evaluation was done every 1000 epochs. We report the NLL approximations obtained on the test set. Best results for the RBM, oRBM and iRBM are reported in Table~\ref{tab:results_caltech}. Again, the oRBM and the iRBM models reach competitive performance compared to the RBM. Samples from all three models are illustrated in Figure~\ref{fig:caltech_samples}.

    The best RBM (500 hidden units) was trained without any regularization and $lr\!=\!10^{-2}$ for 3000 epochs. We used our own implementation to train the RBM. The best oRBM (500 hidden units) was trained with L1 regularization using a regularization factor of $\lambda\!=\!10^{-3}$ and $lr\!=\!10^{-2}$ for 5000 epochs. After 4000 epochs, the best iRBM had 915 hidden units with non-zero weights. It was trained with L1 regularization using a regularization factor of $\lambda\!=\!10^{-3}$ and $lr\!=\!5\!\!\times\!\!10^{-2}$.

    Again, to show that our best iRBM does find an appropriate number of hidden units, we compared it with two others RBMs having respectively 100 and 2000 hidden units. Both were trained without any regularization and respectively with $lr\!=\!10^{-1}$ for 5000 epochs and $lr\!=\!10^{-2}$ for 2000 epochs. Results are reported in Table~\ref{tab:results_caltech} where we can see the oRBM and the iRBM still achieve competitive results compared to the RBM with 2000 hidden units.

\section*{Conclusion}
  We proposed a novel extension of the RBM, the infinite RBM, which obviates the need to specify the hidden layer size. The iRBM is derived from the ordered RBM by taking the infinite limit of its hidden layer size. We presented a training procedure, derived from Contrastive Divergence, such that training the iRBM yields a learning procedure where the effective hidden layer size can grow.

  In future work, we are interested in generalizing the idea of a growing latent representation to structures other than a flat vector representation. We are currently exploring extensions of the RBM allowing for a tree-structured latent representation. We believe a similar construction, involving a similar $z$ random variable, should allow us to derive a training algorithm that also learns the latent representation's size.

\subsection*{Acknowledgments}
  We thank NSERC for supporting this research, Nicolas Le Roux for discussions and comments, and Stanislas Lauly for making the iRBM's training video.

\clearpage
\appendix
\section{Partial derivatives}
  \label{AppendixA}
  \subsection*{Partial derivatives related to the RBM}
    Recall equation~\eqref{rbm:free_energy} representing the free energy of the RBM:
    \[\Frbm(\bv) = -\bv^T\bbvis -\sum_{i=1}^K \softplus(\bW_{i\cdot}\bv + \bhid_i)\]
    Taking the partial derivatives of $\Frbm(\bv)$ \wrt $W_{ij}$, $\bhid_i$ and $\bvis_j$ respectively, we obtain the following:
    \begin{align}
      \label{appendix:rbm:grad_w}
      \frac{\partial \Frbm(\bv)}{\partial W_{ij}}
        &= -\sum_{k=1}^K \frac{\partial\,\softplus(\bW_{k\cdot}\bv + \bhid_k)}{\partial W_{ij}}
        = -\sum_{k=1}^K \sigm(\bW_{k\cdot}\bv + \bhid_k) \frac{\partial\bW_{k\cdot}\bv}{\partial W_{ij}}
        = -\sigm(\bW_{i\cdot}\bv + \bhid_i) v_{j} \\
      \label{appendix:rbm:grad_bhid}
      \frac{\partial \Frbm(\bv)}{\partial \bhid_i}
        &= -\sum_{k=1}^K \frac{\partial\,\softplus(\bW_{k\cdot}\bv + \bhid_k)}{\partial \bhid_i}
        = -\sum_{k=1}^K \sigm(\bW_{k\cdot}\bv + \bhid_k) \frac{\partial\bhid_k}{\partial \bhid_i}
        = -\sigm(\bW_{i\cdot}\bv + \bhid_i) \\
      \label{appendix:rbm:grad_bvis}
      \frac{\partial \Frbm(\bv)}{\partial \bvis_j}
        &= -\frac{\bv^T\bbvis}{\partial \bvis_j}
        = -v_j
    \end{align}
    where $\sigm(\bW_{i\cdot}\bv + \bhid_i)$ can be expressed as a conditional expectation over $h_i$ using equation~\eqref{rbm:ph1_v}
    \begin{align}
      \sigm(\bW_{i\cdot}\bv + \bhid_i)
        = P(h_i=1|\bv)
        = \sum_{h_i\in\{0,1\}} P(h_i=1|\bv)h_i
        = \EE[h_i|\bv] \nonumber
    \end{align}

  \subsection*{Partial derivatives related to the oRBM and the iRBM}
    Recall equation~\eqref{orbm:p_v_free_energy} representing the free energy of the oRBM:
    \[
      \Forbm(\bv) = \ln\left(\sum_{z=1}^K e^{-F(\bv,z)}\right)
    \]
      where
    \[
      \Forbm(\bv,z) = -\bv^T\bbvis -\sum_{i=1}^z \left( \softplus(\bW_{i\cdot}\bv + \bhid_i) - \beta\softplus(\bhid_i) \right)
    \]
    The partial derivatives of $\Frbm(\bv, z)$ \wrt $W_{ij}$, $\bhid_i$ and $\bvis_j$ are similar to equations~\eqref{appendix:rbm:grad_w}, \eqref{appendix:rbm:grad_bhid} and~\eqref{appendix:rbm:grad_bvis} from the RBM and are respectively given by
    \begin{align}
      \label{appendix:orbm:grad_w}
      \frac{\partial \Frbm(\bv,z)}{\partial W_{ij}}
        &= -\sum_{k=1}^z \frac{\partial\,\softplus(\bW_{k\cdot}\bv + \bhid_k)}{\partial W_{ij}} \nonumber\\
        &= -\sum_{k=1}^z \sigm(\bW_{k\cdot}\bv + \bhid_k) \frac{\partial\bW_{k\cdot}\bv}{\partial W_{ij}} \nonumber\\
        &= -H(z-i)\,\sigm(\bW_{i\cdot}\bv + \bhid_i) v_{j} \\
      \label{appendix:orbm:grad_bhid}
      \frac{\partial \Frbm(\bv,z)}{\partial \bhid_i}
        &
        = -\sum_{k=1}^z \frac{\partial\,\softplus(\bW_{k\cdot}\bv + \bhid_k) - \beta\softplus(\bhid_k)}{\partial \bhid_i} \nonumber\\
        &= -\sum_{k=1}^z \left(\sigm(\bW_{k\cdot}\bv + \bhid_k)-\beta\sigm(\bhid_k)\right) \frac{\partial\bhid_k}{\partial \bhid_i} \nonumber\\
        &= -H(z-i)\,\left(\sigm(\bW_{i\cdot}\bv + \bhid_i)-\beta\sigm(\bhid_i)\right)\\
      \label{appendix:orbm:grad_bvis}
      \frac{\partial \Frbm(\bv,z)}{\partial \bvis_j}
        &= -\frac{\bv^T\bbvis}{\partial \bvis_j}
        = -v_j
    \end{align}
    with the Heaviside step function denoted as
    \[
      \label{Heaviside}
      H(n) =
      \begin{cases}
        0, & n < 0\\
        1, & n \ge 0
      \end{cases}\;.
    \]

    Then, the partial derivatives of $\Frbm(\bv)$ \wrt $W_{ij}$, $\bhid_i$ and $\bvis_j$ are obtained respectively as follows:
    \begin{align}
      \frac{\partial \Frbm(\bv)}{\partial W_{ij}}
        \label{appendixA:irbm:derivation_grad_W}
        &= \sum_{z=1}^K \frac{e^{-F(\bv,z)}}{\sum_{z'=1}^K e^{-F(\bv,z')}} \frac{\partial \Frbm(\bv,z)}{\partial W_{ij}} \nonumber \\
        &= -\sum_{z=1}^K P(z|\bv) \, H(z-i) \, \sigm(\bW_{i\cdot}\bv + \bhid_i) v_{j} \nonumber \\
        &= -\sum_{z=1}^K H(z-i) \, P(z|\bv) \, P(h_i=1|\bv, z)v_{j} \nonumber \\
        &= -\sum_{z=1}^K \sum_{h_i\in\{0,1\}} H(z-i)h_i \, P(z|\bv) \, P(h_i|\bv, z)v_{j} \nonumber \\
        &= -\EE_{h_i,z} [H(z-i) h_i | \bv] v_{j} \\
      \frac{\partial \Frbm(\bv)}{\partial \bhid_i}
        \label{appendixA:irbm:derivation_grad_b}
        &= \sum_{z=1}^K P(z|\bv) \frac{\partial \Frbm(\bv,z)}{\partial \bhid_i} \nonumber \\
        &= -\sum_{z=1}^K P(z|\bv) \, H(z-i) \, \left(\sigm(\bW_{i\cdot}\bv + \bhid_i) - \beta\sigm(\bhid_i)\right) \nonumber \\
        &= -\sum_{z=1}^K H(z-i) \, P(z|\bv) \, \left(P(h_i=1|\bv, z) - \beta\sigm(\bhid_i)\right) \nonumber \\
        &= -\sum_{z=1}^K H(z-i) \, P(z|\bv) \, \left(-\beta\sigm(\bhid_i)(1-P(h_i=1|\bv, z) + (1-\beta\sigm(\bhid_i))P(h_i=1|\bv, z) \right) \nonumber \\
        &= -\sum_{z=1}^K H(z-i) \, P(z|\bv) \, \left((0-\beta\sigm(\bhid_i))P(h_i=0|\bv, z) + (1-\beta\sigm(\bhid_i))P(h_i=1|\bv, z) \right) \nonumber \\
        &= -\sum_{z=1}^K \sum_{h_i\in\{0,1\}} H(z-i) \left(h_i - \beta\sigm(\bhid_i)\right) \, P(z|\bv) \, P(h_i|\bv, z) \nonumber \\
        &= -\EE_{h_i,z} \left[H(z-i) \left(h_i - \beta\sigm(\bhid_i)\right) | \bv\right] \\
      \frac{\partial \Frbm(\bv)}{\partial \bvis_j}
        &= -\frac{\bv^T\bbvis}{\partial \bvis_j}
        = -v_j \nonumber
    \end{align}

    Observe that in equations \eqref{appendixA:irbm:derivation_grad_W} and \eqref{appendixA:irbm:derivation_grad_b}, $\sum_{z=1}^K P(z|\bv) \, H(z-i)$ corresponds to $P(z\ge i|\bv)$. This then translates to $(1-\mathbf{cdf}(z|\bv))$ when deriving the gradients as shown in equations \eqref{orbm:grad_W} and \eqref{orbm:grad_b}.

\clearpage
\section{Convergence of the partition function for the iRBM}
  \label{AppendixB}

  We will show that the partition function $Z$ of the iRBM is finite. To do so, we take the limit of $K\rightarrow \infty$ of equation~\eqref{orbm:Z}:
  \begin{align}
    \label{irbm:Z}
    Z &= \sum_{\bv\in\cV} \sum_{z=1}^\infty \sum_{\bh\in\cH_{z}} e^{-\Eorbm(\bv, \bh, z)} \nonumber\\
      &= \sum_{\bv\in\cV} \sum_{z=1}^\infty e^{-\Forbm(\bv, z)} \nonumber\\
      &= \sum_{\bv\in\cV} Z(\bv)
  \end{align}
  Since the sum over all $\bv\in\cV$ is finite and we know from equation~\eqref{irbm:geom} that $Z(\bv)$ is finite, then $Z$ is also finite.

\clearpage

\end{document}